\documentclass{article}

\usepackage{microtype}
\usepackage{graphicx}
\usepackage{subcaption}
\usepackage{booktabs} % for professional tables
\usepackage{tabularx}

\usepackage{hyperref}
\usepackage{float} % Add this to your preamble if not already included

\usepackage[accepted]{icml2025}

\usepackage{amsmath}
\usepackage{amssymb}
\usepackage{mathtools}
\usepackage{amsthm}

\usepackage[capitalize,noabbrev]{cleveref}

\theoremstyle{plain}

\theoremstyle{definition}

\theoremstyle{remark}

\usepackage[textsize=tiny]{todonotes}

\icmltitlerunning{RoboGrasp: A Universal Grasping Policy for Robust Control}

\begin{document}

\twocolumn[
\icmltitle{RoboGrasp: A Universal Grasping Policy for Robust Robotic Control}



\begin{icmlauthorlist}
\icmlauthor{Yiqi Huang}{zcai}
\icmlauthor{Travis Davies}{zcai}
\icmlauthor{Jiahuan Yan}{zcai}
\icmlauthor{Xiang Chen}{pku}
\icmlauthor{Yu Tian}{hav}
\icmlauthor{Luhui Hu}{zcai}
\end{icmlauthorlist}

\icmlaffiliation{zcai}{ZhiCheng AI, Hangzhou, China}
\icmlaffiliation{pku}{Peking University}
\icmlaffiliation{hav}{Harvard University}

\icmlcorrespondingauthor{Yiqi Huang}{yiqi.huang.19@outlook.com}
\icmlcorrespondingauthor{Luhui Hu}{luhuihu@gmail.com}

\icmlkeywords{Machine Learning, Robotics, Diffusion Transformers, Multimodal Computer Vision, Imitation Learning, ICML}

\vskip 0.3in
]



\printAffiliationsAndNotice{\icmlEqualContribution} % otherwise use the standard text.

\begin{abstract}
Imitation learning and world models have shown significant promise in advancing generalizable robotic learning, with robotic grasping remaining a critical challenge for achieving precise manipulation. Existing methods often rely heavily on robot arm state data and RGB images, leading to overfitting to specific object shapes or positions. To address these limitations, we propose RoboGrasp, a universal grasping policy framework that integrates pretrained grasp detection models with robotic learning. By leveraging robust visual guidance from object detection and segmentation tasks, RoboGrasp significantly enhances grasp precision, stability, and generalizability, achieving up to 34\% higher success rates in few-shot learning and grasping box prompt tasks. Built on diffusion-based methods, RoboGrasp is adaptable to various robotic learning paradigms, enabling precise and reliable manipulation across diverse and complex scenarios. This framework represents a scalable and versatile solution for tackling real-world challenges in robotic grasping.
\end{abstract}

\begin{figure*}[ht]
    \centering
    \includegraphics[width=0.9\textwidth]{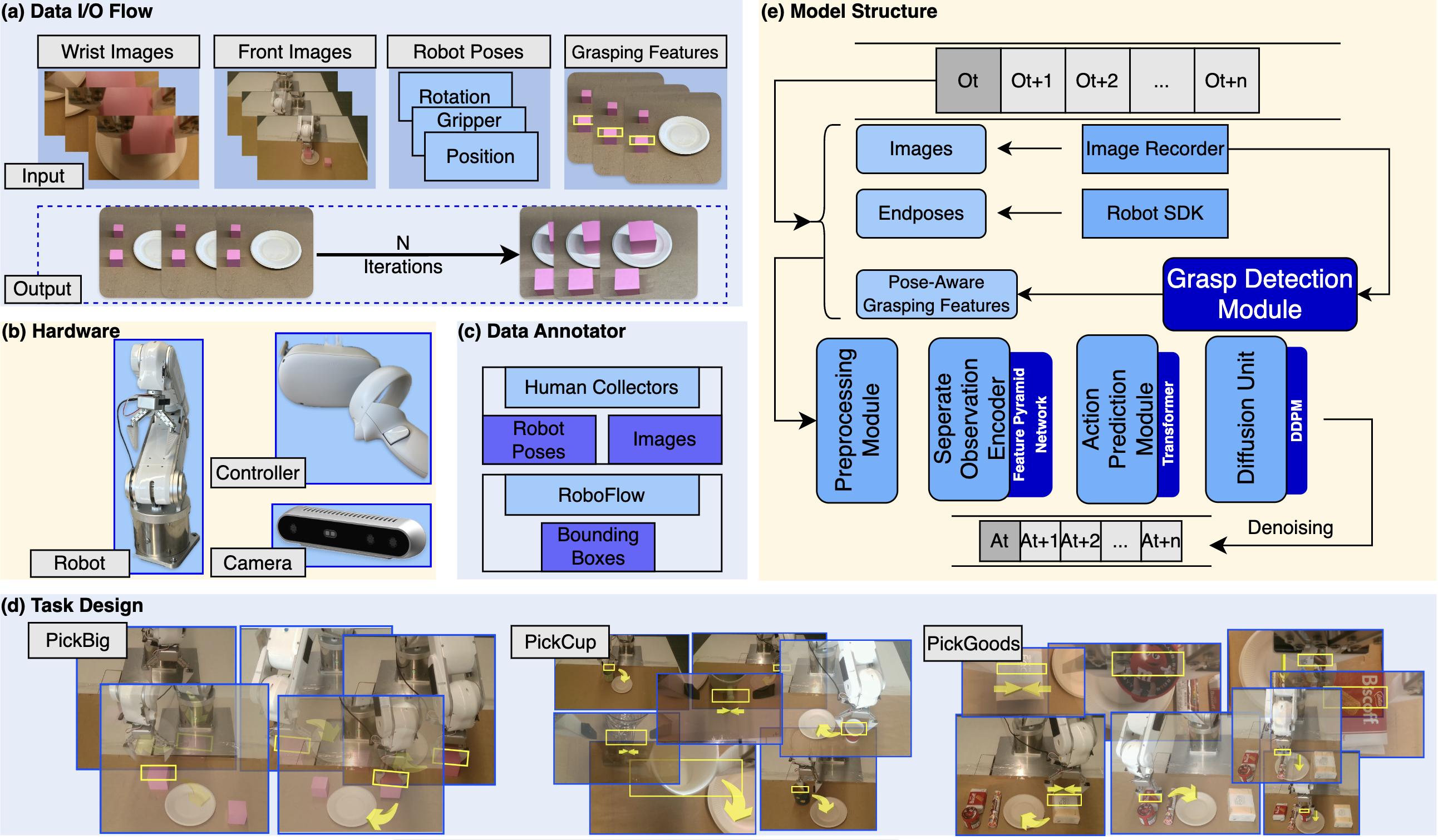}
    \caption{An overview RoboGrasp architecture, demonstrating the integration of grasping guidance, RGB images and robot state data to enhance generalizability and precision of grasping manipulation. (a) Data flow and datasets used for training and inference. (b) Hardware setup, including an industrial-grade robotic arm, RealSense cameras, and a Quest VR headset for data collection. (c) Annotation of demonstrations for grasping affordances. (d) Experimental task designs. (e) The RoboGrasp policy architecture.}
    \label{fig:overview}
\end{figure*}

\section{Introduction}
When a baby encounters an object for the first time, it can often grasp it instinctively. For robots, however, this task is far more complex. Policies trained for one object often fail to generalize to others. Recent advances in Behavior Cloning, particularly diffusion-based policies, have emerged as a promising solution, offering flexibility and expressiveness in handling complex, multi-modal action spaces \cite{diffusion-human-behaviour, diffusion-policy}.

However, Behavior Cloning still face challenges in generalizing beyond their training environments, particularly in dynamic, cluttered settings with unseen or distractor objects. A key limitation lies in their reliance on raw sensor data for conditional input during training and inference \cite{diffusion-policy, 3d-dp}. Without explicit task guidance, these policies depend on implicit patterns learned from data, limiting their robustness \cite{grad-cam}.

To address this, we propose leveraging advancements in computer vision to enhance perception. Pre-trained vision models for tasks like object detection \cite{yolo}, segmentation \cite{sam2}, pose estimation \cite{hr-posenet}, and depth estimation \cite{depth-anything} can provide structured, task-relevant information. By integrating these models, robotic policies can focus on relevant objects and regions, even in cluttered environments, enabling scalable generalization to novel objects and tasks.

We introduce RoboGrasp, a universal grasping policy framework that integrates an auxiliary grasping-box detection model. This model identifies precise grasp regions, providing explicit spatial guidance for the robot arm. By conditioning the policy on these grasping boxes, RoboGrasp enhances generalizability and adaptability.

Our experiments explore two key questions: (1) Can RoboGrasp leverage grasping-based affordances for effective \textbf{few-shot learning} on new or unseen objects? (2) Can it use \textbf{grasping-based affordance prompts} as visual cues to define objectives and generate effective policies? These questions aim to evaluate the scalability of grasping-based affordances for robust, generalizable robot manipulation in real-world environments.

This work represents a step toward deploying robots in unstructured settings, reducing reliance on controlled lab data and improving adaptability for diverse, dynamic tasks.

\section{Related Works}
Recent advancements in robot policy planning have facilitated the democratization of Behavior Cloning (BC), extending its reach beyond specialized research labs \cite{aloha, aloha2, umi}. These approaches typically involve models that map sensor observations into trajectories of future robot poses. In this context, diffusion models have emerged as a powerful tool to address critical limitations of Behavior Cloning, such as covariate shift \cite{covariate-shift}, where robots fail to generalize beyond their training data \cite{domain-generalisation}. Diffusion-based policies, exemplified by Diffusion Policy (DP) \cite{diffusion-policy}, overcome these challenges by generating diverse and multi-modal action trajectories, significantly improving robustness in dynamic and unpredictable environments.

Recent large-scale robotic expert demonstration datasets \cite{x-embodiment} have fueled efforts to scale BC architectures. Works like Robotics Diffusion Transformer (RDT) \cite{rdt}, Octo \cite{octo}, and $\pi_0$ \cite{pi0} demonstrate that skills learned from diverse datasets can transfer to novel tasks, with some models achieving zero-shot generalization to grasping new objects. However, training large-scale models remains computationally expensive, limiting accessibility for resource-constrained settings.

Recent efforts have investigated point-based affordance representations \cite{moka, kalie, rekep}, where keypoints are used to identify task-relevant objects and guide the policy with structured information, often leveraging pre-trained vision models. While scalable, these approaches primarily convey object locations but lack actionable information on how to grasp or manipulate them effectively.

Grasping-based affordance representations offer a more comprehensive solution by encoding feasible grasping strategies \cite{grasping-survey}, providing both spatial and actionable information. Datasets like Grasp Anything \cite{grasp-anything} highlight the potential for scalable data collection in this domain. However, integrating grasping affordances with diffusion-based policies remains underexplored. Existing works such as GQCNN \cite{gqcnn} provide initial steps, but further research is needed to unlock the full potential of affordance-driven planning.

Our work bridges this gap by integrating grasping-based affordance representations with diffusion-based policies. By providing richer conditional inputs, we aim to improve the efficiency and generalization of robot planning models, particularly in resource-constrained settings.

\section{RoboGrasp Policy}
This section outlines the architecture of the RoboGrasp, an augmented variation of Diffusion Policy (DP) designed to incorporate grasp-specific information for improved robotic manipulation. Key enhancements include the integration of a Grasp Detection Module and modifications to the observation encoder. Hyperparameters, such as the number of historical timesteps (2) and predicted actions (16), remain consistent with the original DP framework.

\begin{figure}[H]
    \centering
    \includegraphics[width=0.7\linewidth]{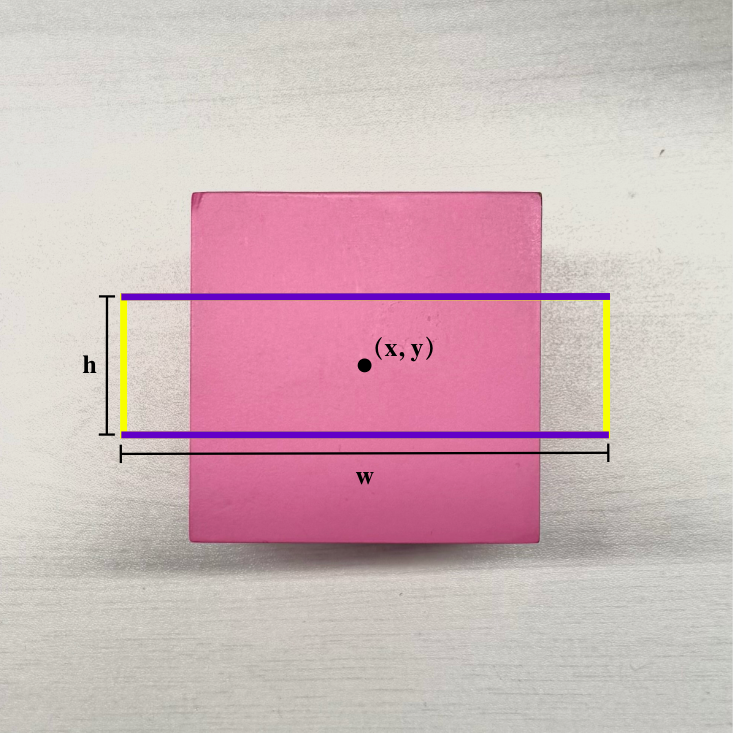}
    \caption{The anatomy of a grasping box. A region on an item indicating the region that can be grasped, along with the $x$, $y$ coordinates of the box's centroid and the box's width and height.}
    \label{fig:grasping_box}
\end{figure}

The grasping box information includes, as shown in Figure \ref{fig:grasping_box}, the $x$ and $y$ coordinate of the grasping box's central point along with the height and width of the box. Normally the angle of rotation in relation to the camera's orientation is also included, however since the robot arm used in this experiment cannot rotate, these parameter was considered redundant in experiments, and all objects were left in unrotated positions.

\subsection{Grasp Detection Module}
The Grasp Detection Module leverages YOLOv11-m \cite{yolo} for its speed, simplicity, and generalizability. YOLOv11-m was fine-tuned on a custom-labeled dataset to predict the object class, 2D spatial coordinates of the grasping box's center, and the box's width and height. During policy training, labels generated by the Grasp Detection Module were directly utilized, while at inference, YOLOv11-m dynamically predicts grasping boxes for the observed data. To simplify grasp selection, the module outputs only the box with the highest confidence score for each run, as the task involves grasping a single object per experiment.

\subsection{Observation Encoder}
The observation encoder combines visual and low-dimensional data into a unified latent representation. A ResNet34-based feature pyramid encoder is employed for each camera view, processing multiview RGB data separately before concatenation. Low-dimensional inputs, such as the robot arm's end pose and gripper sensor data, are incorporated following the original DP design. A novel augmentation introduces grasping box features—class label and spatial information—into the concatenated observation data.

This concatenated data is projected into a fixed-dimensional latent space, serving as a single token per timestep. To capture temporal dependencies, an untrained, lightweight transformer applies self-attention across the designated historical timesteps.

\subsection{Diffusion Action Head}
The action head utilizes a lightweight diffusion transformer, identical to that in DP, to predict actions over 16 timesteps. A DDIM scheduler \cite{iddpm} with a Cosine Beta noise schedule is used for denoising, ensuring efficient and smooth sampling.

Cross-attention mechanisms condition the noised actions on observation tokens, enabling the policy to integrate visual and spatial context effectively. Actions are linearly projected into the latent space for processing within the transformer and are subsequently reprojected into their original dimensions via dedicated linear layers.

\section{Experiments}

Addressing the need for grasp-focused tasks is essential to overcome the limitations inherent in traditional robot learning experiments. Commonly, these experiments utilize similar object targets, allowing models to extensively learn from these specific objects and their associated task completion methods. However, real-world scenarios frequently present a diverse array of objects with varying sizes, types, and grasping requirements, challenging the model’s ability to generalize effectively. To bridge this gap, we designed three primary tasks, \textbf{PickBig}, \textbf{PickCup}, and \textbf{PickGoods} to evaluate the robot’s capability to perform accurate grasping across different object sizes, varied object types with distinct grasping strategies, few-shot learning abilities, and promptable grasping actions.

\begin{figure}[h!]
    \centering
    \includegraphics[width=0.9\linewidth]{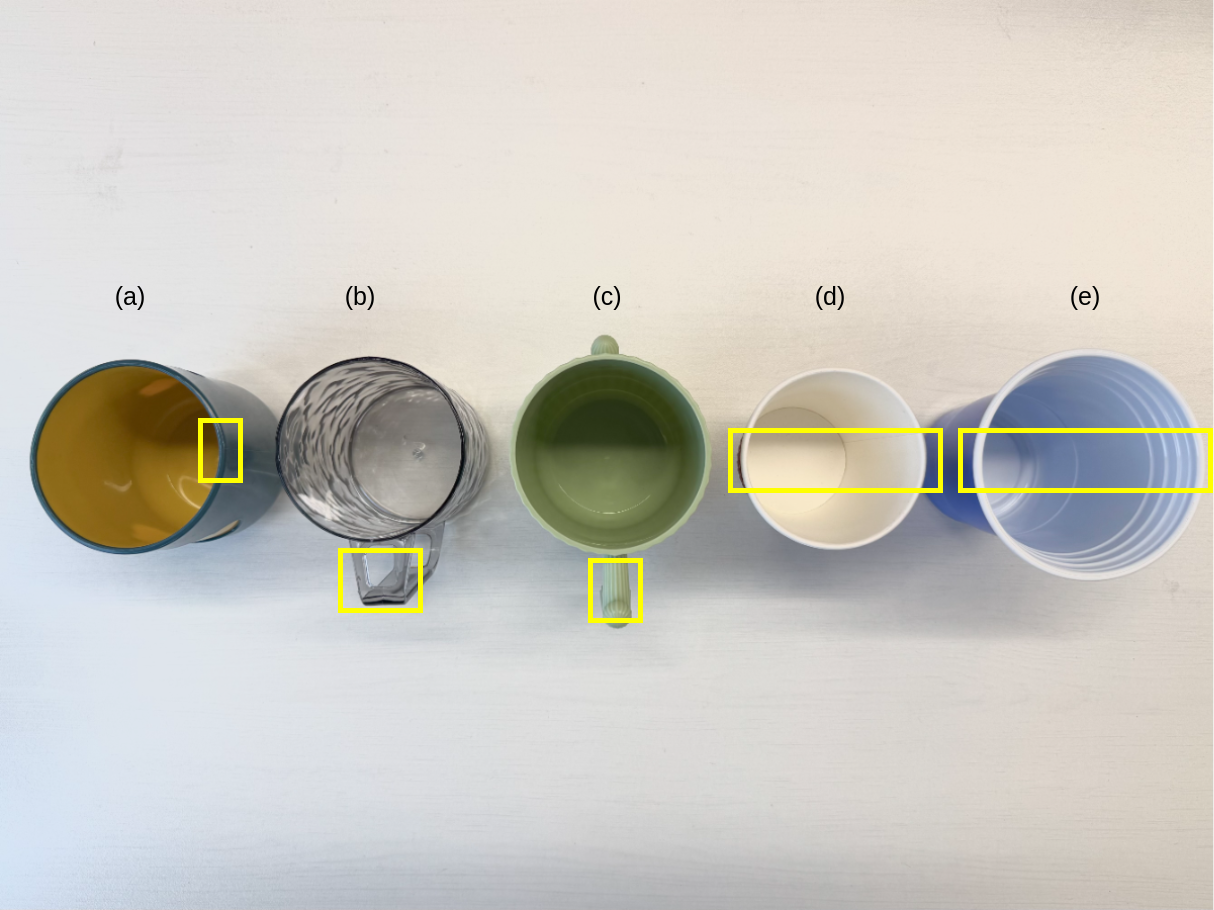}
    \caption{Grasping boxes for cups used in the experiments, shown in a bird's-eye view. (a) illustrates a grasp by the wall of the cup, while (b) and (c) demonstrate grasps by the cup handles. (d) and (e) depict grasps over the cup's diameter. (c) and (e) represent the cups used in the few-shot experiments.}
    \label{fig:cups_grasping}
\end{figure}

\subsection{Task Description}
\begin{figure}[h!]
    \centering
    \includegraphics[width=\linewidth]{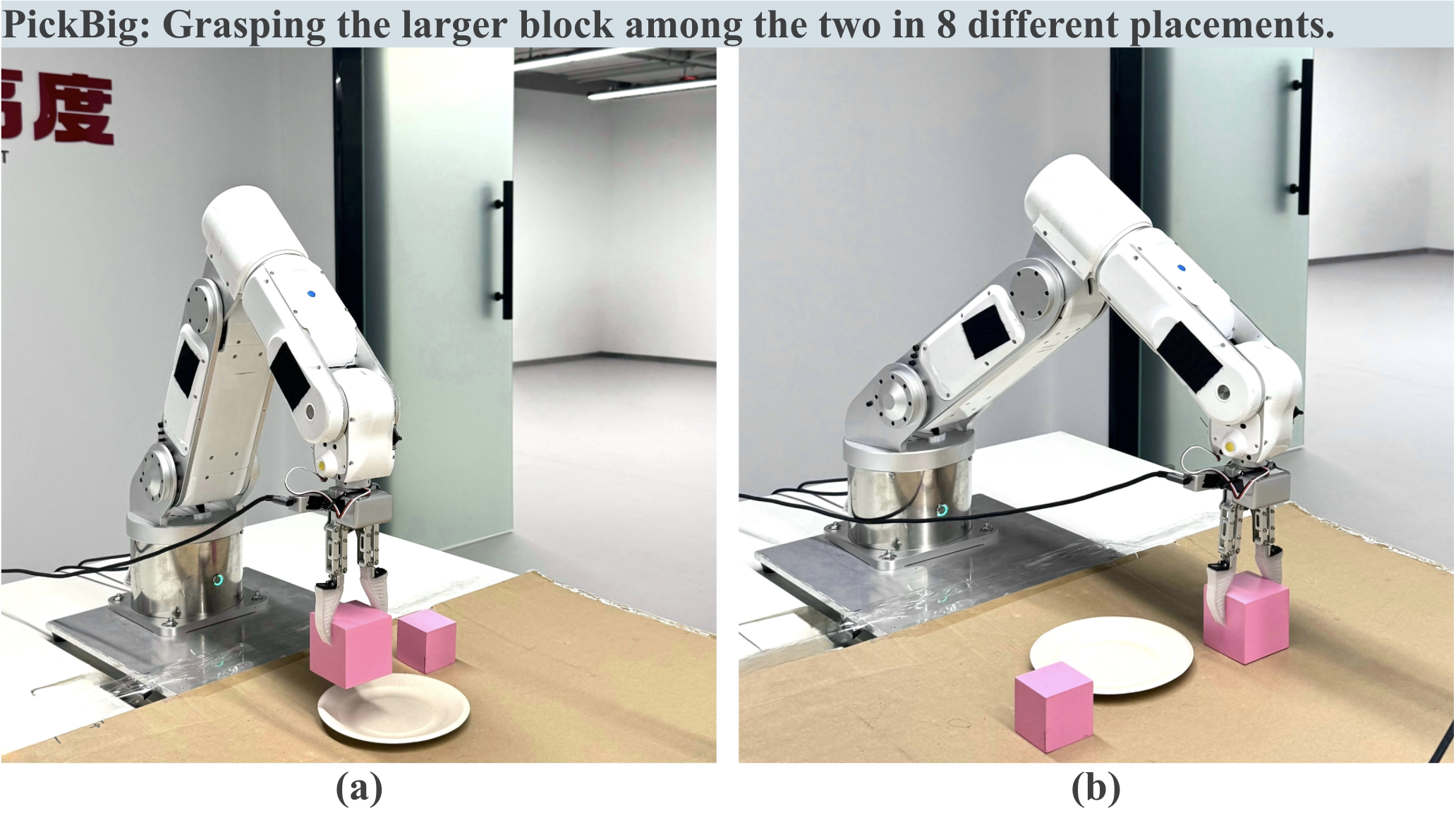}
        \caption{Placement Positions Generalizability experiment setup for PickBig. (a) and (b) show two of the eight placement positions. The objective of PickBig is to distinguish between two similarly shaped blocks and successfully grasp the larger one along its diameter.}
    \label{fig:big_grasping}
\end{figure}

\textbf{PickBig}: 
This task evaluates the robot's ability to distinguish and grasp the larger of two nearly identical blocks, differing only in size. The blocks are placed in eight distinct positions within the workspace, introducing variability in both object dimensions and spatial arrangements. PickBig (see Figure \ref{fig:big_grasping}) assesses the model's capacity to adapt its grasping strategies to accommodate size differences and spatial diversity, ensuring accurate and stable grasps across various scenarios. A key challenge lies in defining the task's goal, and the task aims to test whether providing grasping-based affordance regions helps clarify and achieve the objective more effectively. This focus on goal definition and adaptability makes PickBig a robust test of the model's precision and responsiveness in goal-oriented grasping tasks

\begin{figure}[h!]
    \centering
    \includegraphics[width=\linewidth]{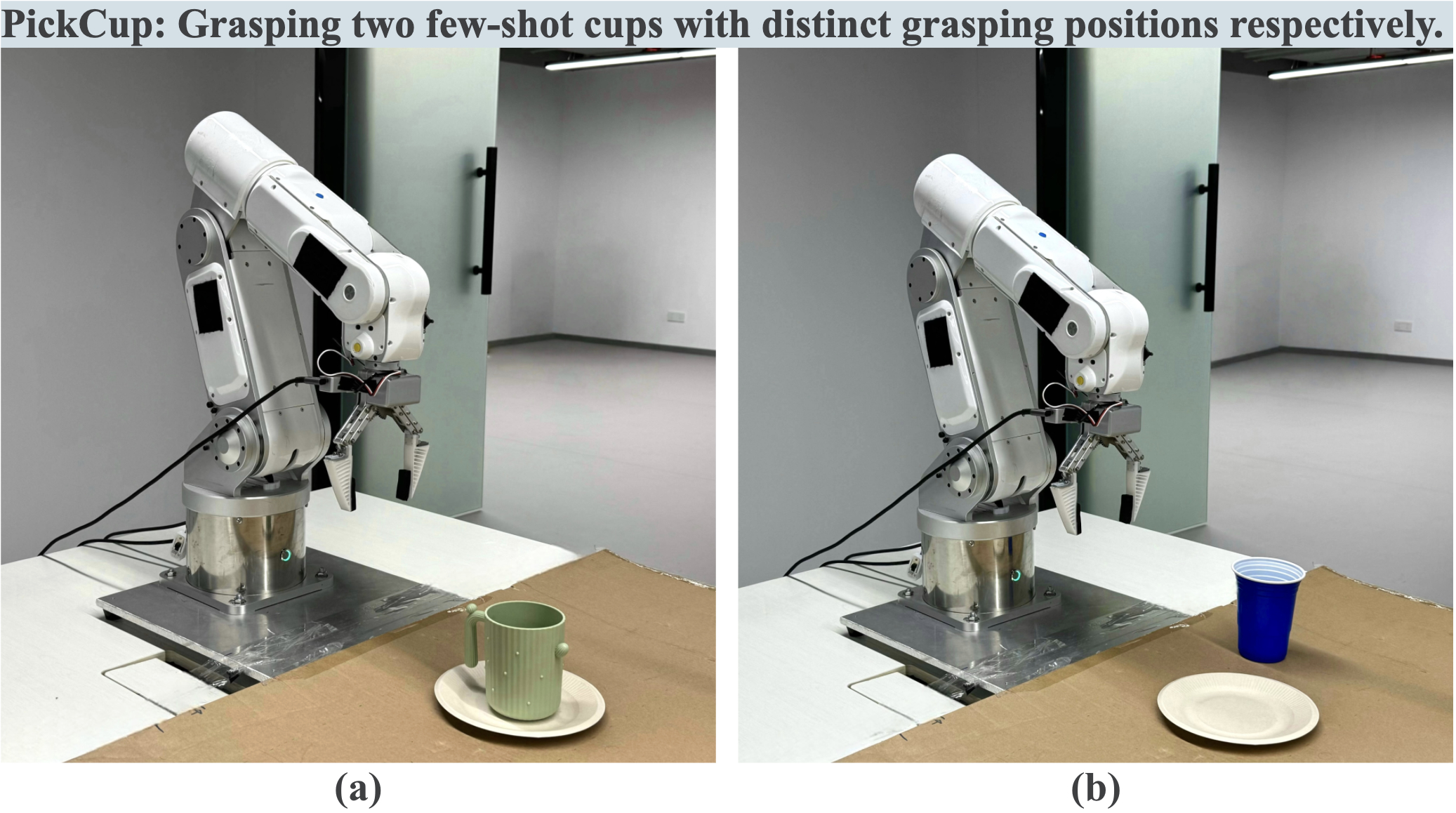}
    \caption{Few-shot experiment setup for PickCup task. The green mug in (a) represents the handle grasping few-shot task with only 5 demonstrations. The blue plastic cup in (b) represents the diameter grasping few-shot task with 10 demonstrations.}
    \label{fig:PickCup_fewshot}
\end{figure}

\textbf{PickCup}: This task focuses on the robot's proficiency in grasping different types of cups using various grasping strategies, as shown in Figure~\ref{fig:cups_grasping}. Three types of cups are employed, each presented with a distinct grasping pattern. Additionally, cups are placed in four distinct positions to introduce variability in grasping scenarios. To evaluate the model's few-shot learning ability, additional instances are introduced with a limited number of demonstration trials (see Figure \ref{fig:PickCup_fewshot}). This inclusion assesses the model's capacity to generalize grasping strategies to new or less-represented objects with minimal training data, ensuring robustness and adaptability in handling diverse and unfamiliar cup types.

\begin{figure}[h!]
    \centering
    \includegraphics[width=\linewidth]{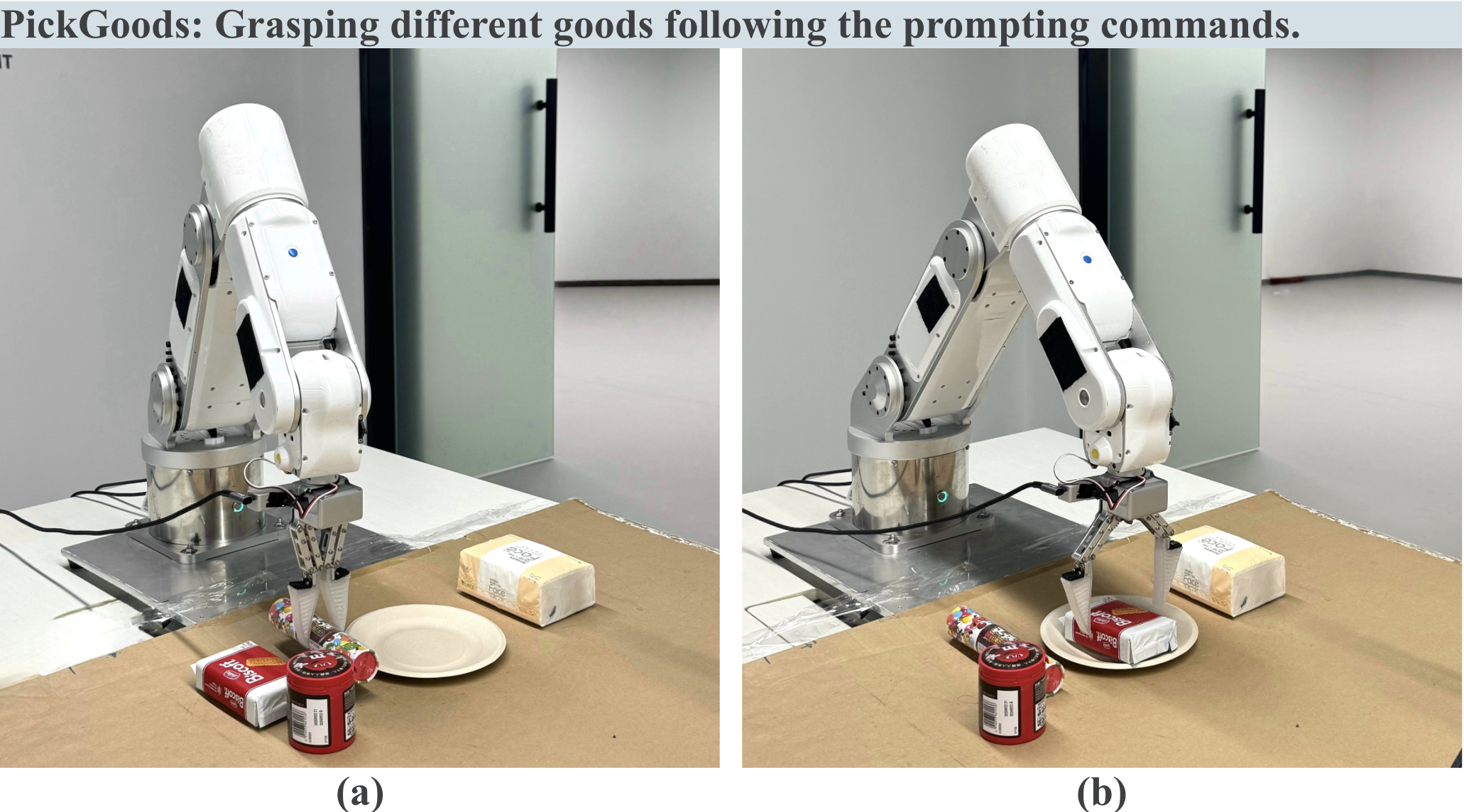}
    \caption{Promptable Grasping experiment setup for the PickGoods task. In (a), the grasping box for the chocolate bar is provided, while in (b), the grasping box for the biscuit is provided. The objective is to follow the grasping box prompts and successfully pick the specified item.}
    \label{fig:goods_grasping}
\end{figure}

\textbf{PickGoods}: This task evaluates the robot's ability to generalize grasping strategies across a wide variety of retail goods, simulating real-world retail environments (see Figure \ref{fig:goods_grasping}). A diverse set of retail items, varying in shape, size, and material, are selected to challenge the robot's adaptability. Each item is grasped using a single, consistent grasping pattern, ensuring uniformity in the approach.

The PickGoods task specifically tests RoboGrasp's ability to generate the correct grasping policy based on a provided \textit{grasping-based affordance region}. This region serves as a \textit{spatial prompt}, guiding the policy toward the desired goal. The prompt acts as a critical test of the policy's responsiveness to predefined objectives, a feature that is notably absent in approaches like DP. Unlike DP, which relies solely on conditional sensor data without explicit goal specification, PickGoods incorporates a clear, goal-oriented prompt, enabling the policy to align its actions with the intended outcome.

\subsection{Data Processing}
The datasets for each task are meticulously curated to reflect a wide range of real-world grasping scenarios:

\begin{itemize}
    \item \textbf{PickBig}: Comprises 600 demonstration trials (8 placement positions $\times$ 75 demo each) involving two blocks, positioned in eight distinct configurations.
    \item \textbf{PickCup}: Consists of 315 demonstration trials (3 cup types $\times$ 4 placement positions $\times$ 25 demo $+$ 15 few-shot demo) involving three types of cups, mug, plastic cup, and paper cup, each subjected to three grasping patterns: handle grasping, cup wall grasping, and diameter grasping. Additionally, 15 extra demonstrations for mugs and paper cups is included to evaluate the model's few-shot learning capabilities.
    \item \textbf{PickGoods}: Contains 400 demonstration trials where each of the retail good has 100 demonstrations.
\end{itemize}

The data preprocessing pipeline ensures high-quality and consistent grasping annotations for each task. A representative subset of approximately 500 frames is manually labeled with grasping boxes, covering all scenarios within the task. These annotations serve as the foundation for training the Grasp Detection Module. Fine-tuning the module on this dataset achieves a mean Average Precision (mAP) exceeding 98\%.

\begin{figure}[h!]
    \centering
    \includegraphics[width=\linewidth]{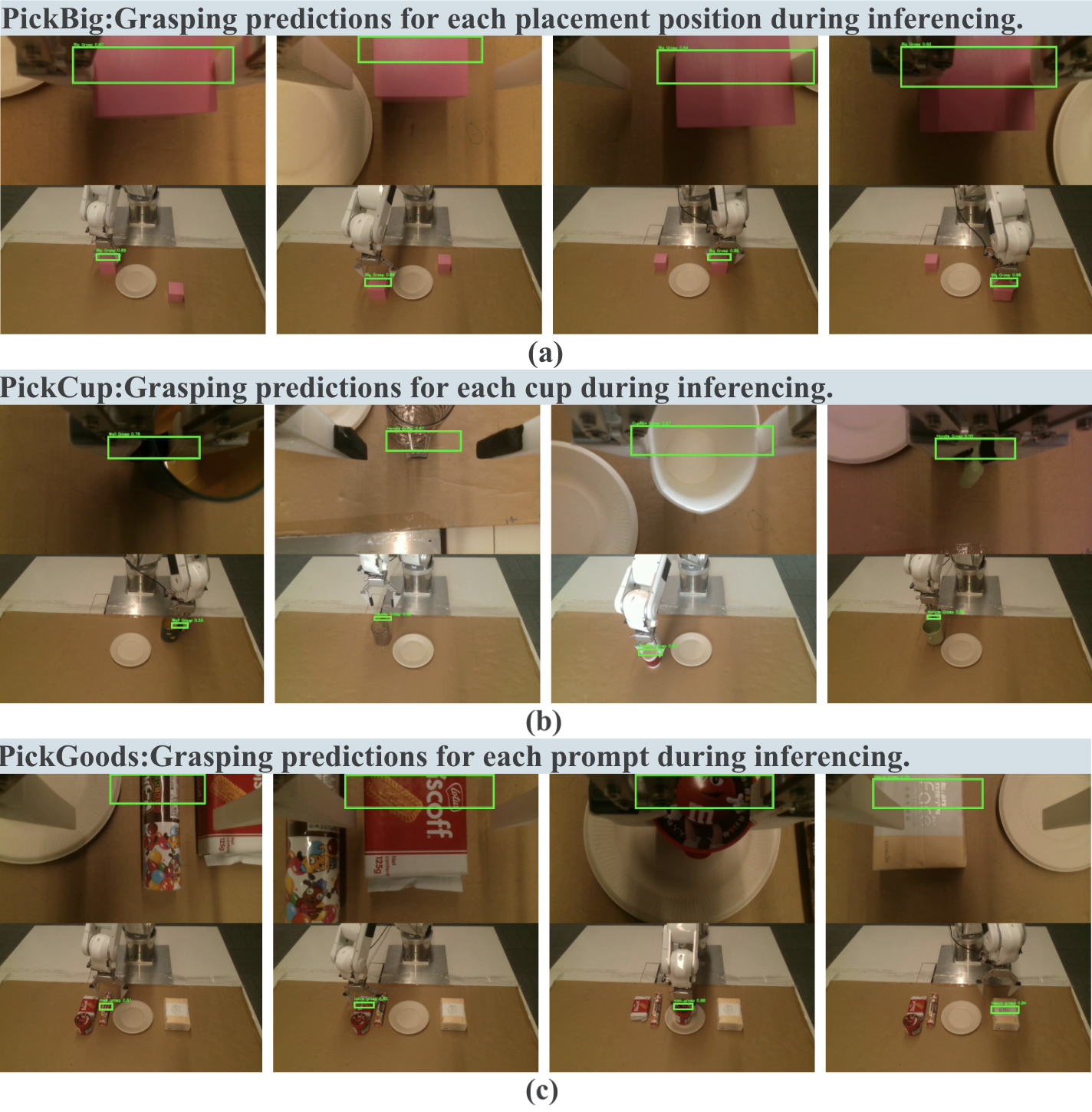}
    \caption{Real-time predictions from our pretrained grasp detection module across different tasks. (a) Demonstrates robust grasping predictions across various placement positions. (b) Highlights accurate detection for diverse grasping strategies. (c) Showcases effective prompting-based predictions.}
    \label{fig:yolo_predict}
\end{figure}

Once trained, the Grasp Detection Module automatically generates grasping boxes for every frame in the collected video data, significantly enhancing annotation scalability while reducing human error. This uniform and reliable annotations support the training of our RoboGrasp policies. During inference, the Grasp Detection Module is seamlessly integrated to predict grasping boxes in real-time (see Figure~\ref{fig:yolo_predict}). This enables the system to dynamically identify optimal grasping regions, ensuring precise and stable grasps across diverse object types and scenarios.

\subsection{Experimental Design and Evaluations}

Although we only use Diffusion Policy (DP) as a baseline, our solution can adapt to various robotic learning frameworks. To analyze how grasping-based affordances enhance model performance, we propose an ablation study across three primary tasks, \textbf{PickBig}, \textbf{PickCup}, and \textbf{PickGoods} to compare two model configurations:

\begin{enumerate}
    \item \textbf{DP Model (Baseline)}: The standard diffusion-based visuomotor policy without any additional enhancements. It serves as a benchmark for measuring the impact of subsequent modifications.
    
    \item \textbf{RoboGrasp Model}: Integrates grasping box annotations generated by the Grasp Detection Module to guide the diffusion process. This focuses the model on optimal grasping regions, improving accuracy and stability.
\end{enumerate}

\subsubsection{Testing Standardization}

To ensure a fair and reliable comparison between DP and RoboGrasp, strict standardization was maintained during inference testing. For each task, the placement positions and involved objects were identical across both models. This controlled setup eliminates variability in testing conditions, ensuring that any performance differences observed are attributable to the models themselves rather than inconsistencies in the experimental setup. Such ablation-based evaluations provide a robust framework for identifying and quantifying improvements in model performance.

\begin{table*}[ht]
\caption{Detailed Performance Comparison Across Tasks for Diffusion Policy and RoboGrasp. The table highlights task success rate and grasping success rate for each target object.}
\label{sample-table}
\vskip 0.15in
\begin{center}
\begin{small}
\begin{sc}
\begin{tabular}{p{1.5cm} p{1cm} l p{0.9cm} p{1.1cm} p{1.6cm} l l l}
\toprule
\textbf{Task} & \textbf{Obj. Num} & \textbf{Target} & \textbf{Demos} & \textbf{Position. Num} & \textbf{Grasping Strategy} & \textbf{Model} & \textbf{TSR(\%)} & \textbf{GSR (\%)} \\
\midrule
PickBig & 2 & Bigger Block & 600 & 8 & Diameter & DP & 67.5 & 66.25 \\
        &   &              &     &   &                   & \textbf{RoboGrasp} & \textbf{97.5} & \textbf{96.25} \\
\cmidrule{1-9}
PickCup & 5 & Grey Mug & 100 & 4 & Handle & DP & 95 & 80 \\
        &   &         &     &   &                 & \textbf{RoboGrasp} & \textbf{100} & \textbf{100} \\
\cmidrule{3-9}
        &   & Blue Plastic Cup & 100 & 4 & Wall & DP & 87.5 & 85 \\
        &   &                 &     &   &             & \textbf{RoboGrasp} & \textbf{92.5} & \textbf{92.5} \\
\cmidrule{3-9}
        &   & Red Paper Cup & 100 & 4 & Diameter & DP & 95 & 70 \\
        &   &              &     &   &                 & \textbf{RoboGrasp} & \textbf{100} & \textbf{100} \\
\cmidrule{3-9}
        &   & Blue Plastic Cup & 10 & 2 & Diameter & DP & 60 & 55 \\
        &   &                 &    &    &               & \textbf{RoboGrasp} & \textbf{100} & \textbf{95} \\
\cmidrule{3-9}
         &   & Green Mug & 5 & 1 & Handle & DP & 70 & 60 \\
        &   &           &   &   &                 & \textbf{RoboGrasp} & \textbf{100} & \textbf{100} \\
\cmidrule{1-9}
PickGoods & 4 & Meiji Chocolate Bar & 100 & 1 & Diameter & DP & 0 & 0 \\
          &   &                     &     &   &                   & \textbf{RoboGrasp} & \textbf{100} & \textbf{98} \\
\cmidrule{3-9}
          &   & Lotus Biscuit & 100 & 1 & Diameter & DP & 0 & 0 \\
          &   &              &     &   &                   & \textbf{RoboGrasp} & 0 & 0 \\
\cmidrule{3-9}
          &   & m\&m & 100 & 1 & Diameter & DP & 0 & 0 \\
          &   &      &     &   &                   & \textbf{RoboGrasp} & \textbf{4} & \textbf{4} \\
\cmidrule{3-9}
          &   & Tissue & 100 & 1 & Diameter & DP & 89.5 & 86 \\
          &   &        &     &   &                   & \textbf{RoboGrasp} & \textbf{100} & \textbf{100} \\
\bottomrule
\end{tabular}
\end{sc}
\end{small}
\end{center}
\vskip -0.1in
\end{table*}

\subsubsection{Evaluation Metrics}

To quantitatively compare the two models, we employed the following metrics:
\begin{enumerate}
    \item \textbf{Task Success Rate (TSR)}: The percentage of successful task completions across all tasks.
    \item \textbf{Grasp Success Rate (GSR)}: Evaluates the effectiveness of grasping strategies by measuring action accuracy, consistency, and stability across diverse objects and scenarios. This metric is particularly useful for assessing the model’s ability to adapt and apply appropriate grasping strategies in varying contexts. GSR is defined as:
\end{enumerate}

\begin{equation}
GSR = \frac{\text{No. Successful Grasps}}{\text{No. Total Grasp Attempts}}
\end{equation}

These metrics provide a comprehensive assessment of each model’s performance, capturing both high-level task success and fine-grained grasping proficiency.

\section{Results and Discussion}

Our experimental results clearly demonstrate RoboGrasp’s superiority over DP across all evaluated tasks. Table~\ref{sample-table} provides a detailed comparison of TSR and GSR, offering insights into how RoboGrasp consistently outperforms DP across various objects, grasping strategies, and placement positions.

In the PickBig task, RoboGrasp achieves a task success rate of 97.5\% and a grasp success rate of 96.25\%, significantly outperforming DP’s 67.5\% and 66.25\%. The effectiveness of grasping box detections is evident as RoboGrasp excels in distinguishing the larger block across eight varied placement positions, adapting seamlessly to positional changes and size differences.

In the PickCup task, involving five distinct objects and diverse grasping strategies, RoboGrasp consistently achieves near-perfect performance, with task success rates of 100\% and grasp success rates ranging from 92.5\% to 100\%. This significantly surpasses DP, which struggles with inconsistent strategies, particularly for challenging cases like blue plastic cups and green mugs. RoboGrasp’s ability to transfer learned skills to few-shot objects highlights its adaptability.

For the PickGoods task, RoboGrasp’s prompt-based grasping successfully handles two out of four target objects, including chocolate bars and tissue packs. While DP struggles with object identification and inconsistent grasps, RoboGrasp utilizes grasping box prompts to focus on the target, achieving up to 100\% task success and 98\% grasp success rates for these objects, demonstrating adaptability despite the task’s challenges.

These results highlight the critical role of grasping box detections in improving task performance and generalization capabilities. RoboGrasp consistently delivers higher task success and grasping success rates across diverse objects, placements, and strategies, demonstrating its adaptability and robustness in various manipulation challenges.

\subsection{Data Compensation}
The size and diversity of the dataset play a critical role in the training and performance of robotic learning models like DP and RoboGrasp. We identified several key factors influencing training and generalization:

\textbf{State Space vs. Number of Demonstrations:} In the PickBig task, the initial dataset comprised only 300 demonstrations. However, training results revealed a high variance in training loss and a large mean squared error (MSE) in action predictions for both DP and RoboGrasp. This inconsistency is likely due to the large state space introduced by eight distinct placement positions, which demands more data to adequately capture the variations. To address this, the dataset size was doubled to 600 demonstrations (75 per placement position), resolving the variance issue and improving training stability.

\textbf{Similar vs. Distinct Objects:} In contrast, the PickCups task demonstrated a different data requirement. For this task, only 25 demonstrations were collected for each cup at each placement position, and the models still achieved convergence during training. This outcome is likely because the cups in this task are distinct, making it easier for the model to differentiate between objects. 

These results suggest that datasets need to be scaled up proportionally to the size of the state space and when the target objects are similar in appearance. The distinction between objects significantly impacts the model’s ability to generalize and perform effectively with fewer demonstrations.

\subsection{Task Performance Analysis}
This section systematically analyzes RoboGrasp’s capabilities and limitations across key dimensions critical to real-world deployment: (1) the interplay between state space complexity and dataset scale in policy optimization, (2) generalization capacity through few-shot skill transfer to novel objects, and (3) responsiveness to spatially grounded affordance cues for deriving context-aware policies. By dissecting performance variations across tasks, we identify how environmental constraints, object distribution sparsity, and affordance grounding collectively shape the system’s adaptability and failure modes.

\subsubsection{Addressing Positioning and State Spaces}
The PickBig task, with its eight distinct placement positions, requires a robust policy to handle a large state space. Increasing demonstrations from 300 to 600 resolved training instability, enabling better state coverage and a fairer comparison between DP and RoboGrasp.

DP struggles with overfitting to low-dimensional robot state representations, leading to repetitive, fixed movements and inconsistent performance. It often fails to adapt to positional changes or differentiate between similar-shaped blocks, limiting its effectiveness.

\begin{figure}[h!]
    \centering
    \includegraphics[width=\linewidth]{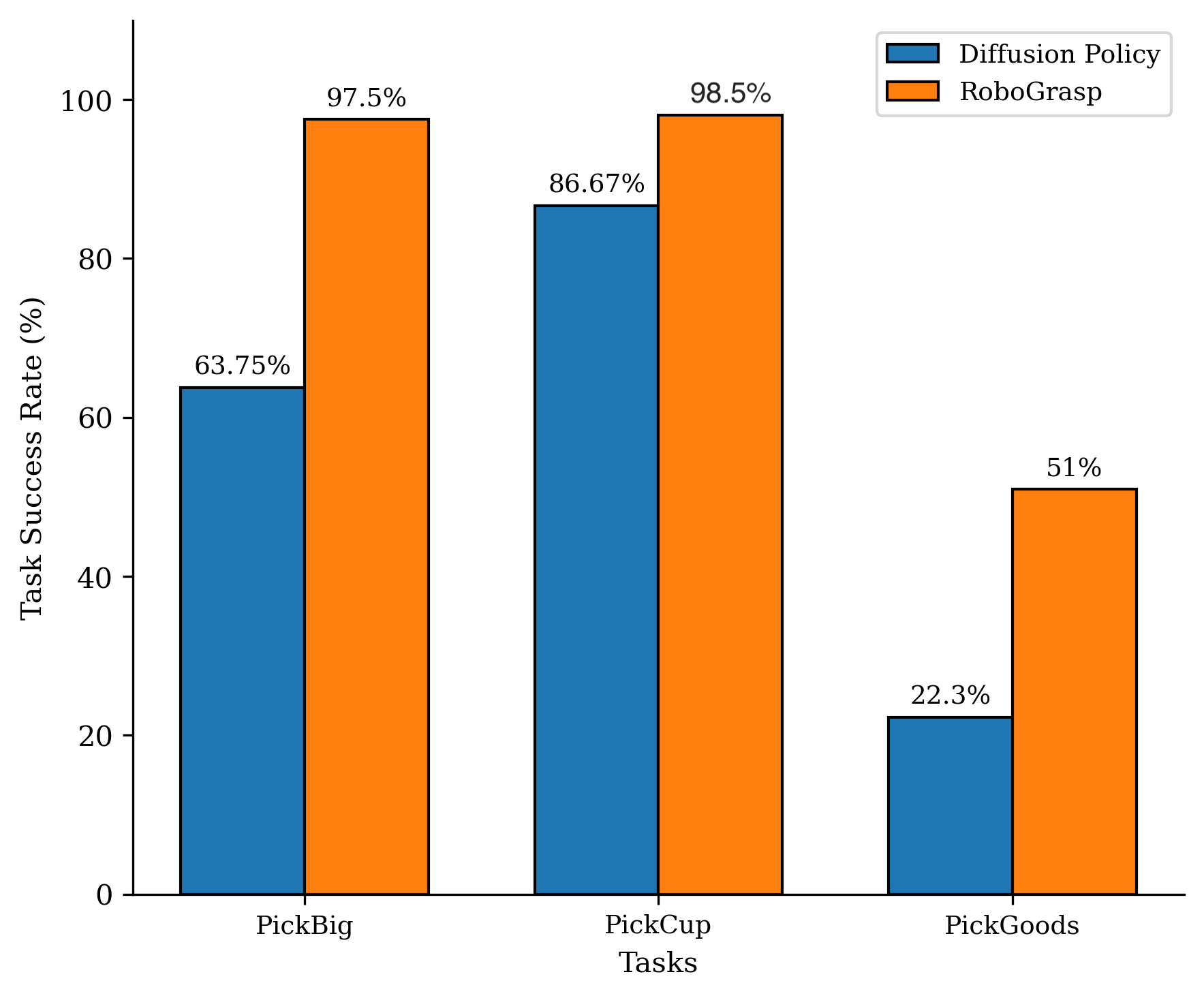} % Adjust width as needed
    \caption{Comparison of average Task Success Rate for DP and RoboGrasp.}
    \label{fig:comparison_tsr}
\end{figure}

In contrast, RoboGrasp achieves a 33.75\% higher TSR by leveraging grasping box detections as additional input (see Figure~\ref{fig:comparison_tsr}). These predictions explicitly guide RoboGrasp to target the correct block, enabling dynamic adaptation to positional changes and precise block differentiation. This results in significantly improved grasping accuracy and consistency.

The PickBig results highlight the importance of state space diversity in training and the critical role of grasping box detections in guiding robotic policies. RoboGrasp’s superior performance demonstrates its potential for generalizable and robust robotic manipulation

\subsubsection{Few-Shot Learning and Strategy Selection}
\begin{figure}[h!]
    \centering
    \includegraphics[width=\linewidth]{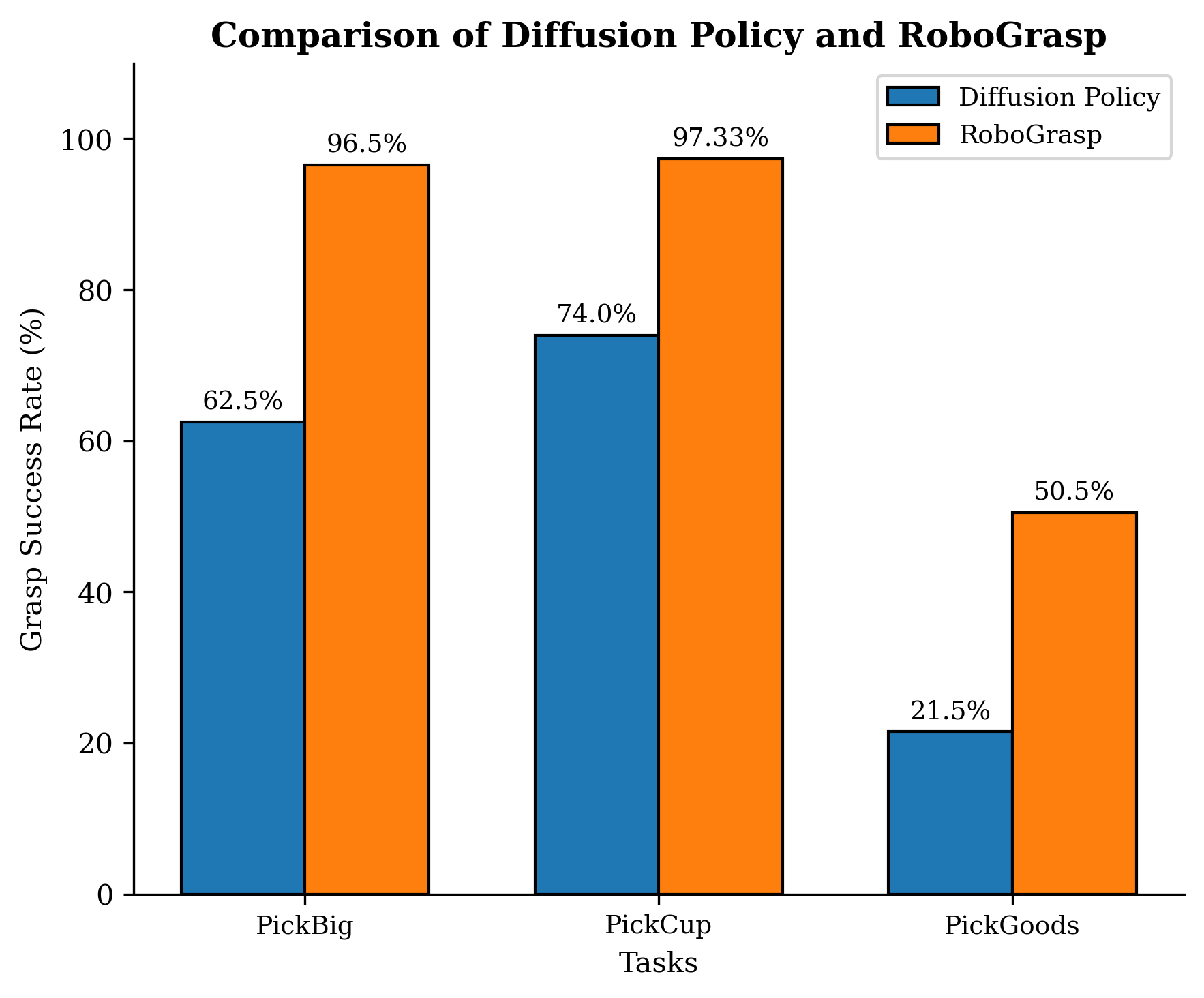} % Adjust width as needed
    \caption{Comparison of average Grasp Success Rate for DP and RoboGrasp.}
    \label{fig:comparison_gsr}
\end{figure}

The PickCups task, with its four placement positions and distinct cup shapes, presents a smaller state space than PickBig, allowing DP to perform better by relying on a narrower diversity of policy action trajectories. However, RoboGrasp outperforms DP with a 23.33\% higher GSR (see Figure~\ref{fig:comparison_gsr}. While DP can complete the pick-and-place task, it often employs inconsistent and mixed grasping strategies, such as using a wall grasp for one cup and a rim grasp for another, failing to generalize effectively. In contrast, RoboGrasp excels in few-shot learning, consistently selecting the correct grasping strategy for cups with only 5 or 10 demonstrations, even when their shapes differ from the primary training set. By leveraging grasping box predictions, RoboGrasp ensures precise and consistent manipulation, demonstrating its ability to generalize across geometric variations with minimal data and adapt to diverse tasks reliably

\subsubsection{Grasping-based affordance Prompt}
The superior performance of RoboGrasp over DP in the PickBig task highlights the efficacy of grasping-based affordance prompts in guiding policy decisions for goal-oriented manipulation. In contrast, while RoboGrasp still significantly outperforms DP in the PickGoods task (28.7\% improvement in TSR), its relative performance decline in this scenario stems from the inherent complexity of the environment: three closely adjacent candidate objects in a confined workspace created a multimodal action probability distribution. This complexity led the policy to prioritize proximal targets (e.g., grasping the first visible item) rather than strategically selecting optimal objects. Notably, the grasping-based affordances did still influence directional choices, such as prompting leftward motions for tissue retrieval or rightward motions for food items. The simpler PickBig environment—with only two candidate objects—resulted in a unimodal action distribution, enabling more deterministic and effective policy execution. However, the fixed spatial placement of objects in PickGoods raises concerns about potential over-reliance on robot pose priors rather than affordance-driven reasoning. To address these limitations, future work could enhance policy robustness by training on datasets with greater positional diversity, thereby reducing environmental bias and improving generalization to cluttered configurations

\section{Conclusion}
RoboGrasp presents a novel approach to robotic grasping, addressing key challenges in precision and generalization by leveraging pretrained grasp detection models. By moving beyond the limitations of traditional reliance on robot state data and RGB images, RoboGrasp enables robust and adaptable grasping across diverse scenarios, demonstrating significant improvements in grasp success rates.

Our experiments evaluated RoboGrasp's ability to transfer grasping skills to new objects through few-shot learning and its capacity to utilize grasping boxes as visual prompts to define goals and generate effective robot policies. The results highlight RoboGrasp's strong few-shot learning capabilities, achieving superior generalization to unseen items compared to DP. Furthermore, RoboGrasp exhibited a substantial performance boost in defining task objectives and generating policies with grasping box prompts, demonstrating its ability to outperform DP by a wide margin.

Built on diffusion-based methods, RoboGrasp is a flexible framework with the potential to scale across a variety of complex manipulation tasks. This work lays a strong foundation for the development of scalable, reliable, and generalizable robotic systems capable of addressing real-world challenges in dynamic and unstructured environments.

\section{Future Directions}
This work highlights several unexplored avenues with significant potential to advance robotic learning and grasping. Language prompting, inspired by methods such as Grounding DINO \cite{grouding_dino} and DINO-X \cite{dinox}, remains underexplored in robotics. Integrating language commands to generate grasping boxes or guide manipulation tasks could greatly enhance flexibility and generalizability. 

Similarly, employing grasp-guided approaches in other frameworks, such as ACT \cite{aloha}, or large foundation models like Robotics Diffusion Transformer (RDT) \cite{rdt}, presents an opportunity to scale robotic learning to broader tasks. Furthermore, incorporating grasping prompts into world models, akin to visual prompts in \cite{motion_prompting}, could enhance real-world modeling and planning, boosting performance in dynamic and unstructured environments. 

Additionally, evaluating these methods in simulation environments would provide a cost-effective way to assess robustness and scalability. Future work could also explore extending grasping capabilities by incorporating rotational parameters into grasping boxes and employing rotatable arms, enabling robots to handle more complex and precise manipulation tasks. These directions offer immense potential to improve precision, generalization, and adaptability in robotics.

\nocite{langley00}

\bibliography{RoboGrasp}
\bibliographystyle{icml2025}

%%%%%%%%%%%%%%%%%%%%%%%%%%%%%%%%%%%%%%%%%%%%%%%%%%%%%%%%%%%%%%%%%%%%%%%%%%%%%%%
%%%%%%%%%%%%%%%%%%%%%%%%%%%%%%%%%%%%%%%%%%%%%%%%%%%%%%%%%%%%%%%%%%%%%%%%%%%%%%%
% APPENDIX
%%%%%%%%%%%%%%%%%%%%%%%%%%%%%%%%%%%%%%%%%%%%%%%%%%%%%%%%%%%%%%%%%%%%%%%%%%%%%%%
%%%%%%%%%%%%%%%%%%%%%%%%%%%%%%%%%%%%%%%%%%%%%%%%%%%%%%%%%%%%%%%%%%%%%%%%%%%%%%%
\newpage

\onecolumn

\end{document}